\relax

\documentclass{llncs}
\usepackage{graphicx}

\usepackage{xcolor}
\usepackage[hidelinks,colorlinks = true,
linkcolor = blue,
urlcolor  = blue,
citecolor = blue,
anchorcolor = blue]{hyperref}
\usepackage{microtype} %
\usepackage[cmex10]{amsmath}
\usepackage{fnbreak} %
\usepackage{url}
\usepackage{longtable}
\usepackage{bm,amssymb}
\usepackage{graphicx}
\usepackage{graphics, color}
\usepackage{algorithm}
\usepackage{algpseudocode}
\usepackage{longtable}
\usepackage{tabularx}
\usepackage{url}
\newfloat{algorithm}{t}{lop}

\usepackage[colorinlistoftodos,prependcaption,textsize=tiny]{todonotes}
\usepackage{enumitem}
\usepackage{thmtools, thm-restate}
\usepackage{xcolor}
\usepackage{subfig}

\usepackage{tcolorbox}

\algblockdefx{MRepeat}{EndRepeat}{\textbf{repeat}}{}
\algnotext{EndRepeat}
\algnotext{EndIf}
\algnewcommand{\algorithmicgoto}{\textbf{go to line}}%
\algnewcommand{\Goto}[1]{\algorithmicgoto~\ref{#1}}%

\algnotext{EndFor}
\algnotext{EndIf}
\algnotext{EndWhile}
\algnotext{EndFunction}
\algblockdefx{Do}{EndDo}{\textbf{do}\;}{}
\algnotext{EndDo}

\newcommand{\integratedImplicit}{\ensuremath{\mathsf{IntegratedImplicit}}}
\newcommand{\greedyIndSearch}{\ensuremath{\mathsf{GreedyIndSearch}}}

\newcommand{\explicit}{\ensuremath{\mathsf{ExplicitSearch}}}

\newcommand{\prob}{\ensuremath{\mathsf{Pr}}}

\newcommand{\satisfying}[1]{\ensuremath{sol({#1})}} %
\newcommand{\ProjectSatisfying}[2]{\ensuremath{sol({#1})}_{\downarrow #2}}

\newcommand{\UniGen}{\ensuremath{\mathsf{UniGen}}}

\newcommand{\PAC}{\ensuremath{\mathsf{PAC}}}

\newcommand{\ApproxCount}{\ensuremath{\mathsf{ApproxCount}}}

\newcommand{\ApproxMCThree}{\ensuremath{\mathsf{ApproxMC3}}}
\newcommand{\ApproxMCFour}{\ensuremath{\mathsf{ApproxMC4}}}
\newcommand{\UniGenThree}{\ensuremath{\mathsf{UniGen3}}}

\newcommand{\UniGenTwo}{\ensuremath{\mathsf{UniGen2}}}
\newcommand{\BplusE}{\ensuremath{\mathsf{B+E}}}
\newcommand{\MIS}{\ensuremath{\mathsf{MIS}}}

\newcommand{\ApproxMC}{\ensuremath{\mathsf{ApproxMC}}}

\newcommand{\Break}{\State \textbf{break} }
\newcommand{\Continue}{\State \textbf{continue} }

\newcommand{\arjun}{\ensuremath{\mathsf{Arjun}}}
\newcommand{\simpleSearch}{\ensuremath{\mathsf{SimpleSearch}}}

\newcommand{\unknown}{\ensuremath{\mathsf{unknown}}}
\newcommand{\indep}{\ensuremath{\mathcal{I}}}

\newcommand{\assumps}{\ensuremath{\mathsf{assumps}}}
\newcommand{\solve}{\ensuremath{\mathsf{solve}}}
\newcommand{\CreateNewVars}{\ensuremath{\mathsf{CreateNewVars}}}

\usepackage{lipsum,atbegshi,etoolbox}%
\makeatletter
\newcommand{\discardpages}[1]{%
	\xdef\discard@pages{#1}%
	\AtBeginShipout{%
		\renewcommand*{\do}[1]{%
			\ifnum\value{page}=##1\relax%
			\AtBeginShipoutDiscard%
			\gdef\do####1{}%
			\fi%
		}%
		\expandafter\docsvlist\expandafter{\discard@pages}%
	}%
}
\newif\ifkeeppage
\newcommand{\keeppages}[1]{%
	\xdef\keep@pages{#1}%
	\AtBeginShipout{%
		\keeppagefalse%
		\renewcommand*{\do}[1]{%
			\ifnum\value{page}=##1\relax%
			\keeppagetrue%
			\gdef\do####1{}%
			\fi%
		}%
		\expandafter\docsvlist\expandafter{\keep@pages}%
		\ifkeeppage\else\AtBeginShipoutDiscard\fi%
	}%
}
\makeatother

\algnotext{EndFor}
\algnotext{EndIf}
\algnotext{EndWhile}
\algnotext{EndFunction}
\algblockdefx{Do}{EndDo}{\textbf{do}\;}{}
\algnotext{EndDo}
\usepackage{booktabs} %

\usepackage[utf8]{inputenc} %

\usepackage{graphicx}
\usepackage{epsfig}
\usepackage{xcolor}
\usepackage{microtype}

\title{
Arjun: An Efficient Independent Support Computation Technique and its Applications to Counting and Sampling\thanks{The tool is available open-source at \url{https://github.com/meelgroup/arjun}}
}
\author{Mate Soos \and Kuldeep S. Meel}
\institute{National University of Singapore}
\begin{document}
\maketitle
\begin{abstract}
	Given a Boolean formula $\varphi$ over the set of variables $X$ and a projection set $\mathcal{P} \subseteq X$, a subset of variables $\mathcal{I}$ is independent support of $\mathcal{P}$ if two solutions agree on $\mathcal{I}$, then they also agree on $\mathcal{P}$. The notion of independent support is related to the classical notion of definability dating back to 1901, and have been studied over the decades. Recently, the computational problem of determining independent support for a given formula has attained importance owing to the crucial importance of independent support for hashing-based counting and sampling techniques.

	In this paper, we design an efficient and scalable independent support computation technique that can handle formulas arising from real-world benchmarks. Our algorithmic framework, called {\arjun}, employs implicit and explicit definability notions, and is based on a tight integration of gate-identification techniques and assumption-based framework. We demonstrate that augmenting the state of the art model counter {\ApproxMCFour} and sampler {\UniGenThree} with {\arjun} leads to significant performance improvements. In particular, {\ApproxMCFour} augmented with {\arjun} counts 387 more benchmarks out of 1896 while {\UniGenThree} augmented with {\arjun} samples 319 more benchmarks within the same time limit.
\end{abstract}
\section{Introduction}
Given a Boolean formula $\varphi$ over the set of variables $X$, let $\satisfying{\varphi}$ represent the set of solutions of $\varphi$. For a given assignment $\sigma$ over $X$ and a subset of variables $\mathcal{P} \subseteq X$, let $\sigma_{\downarrow \mathcal{P}} $ represent the assignment of variables restricted to $\mathcal{P}$.
Given a Boolean formula $\varphi$ over the set of variables $X$ and a projection set $\mathcal{P} \subseteq X$, a subset of variables $\mathcal{I}$ such that $\mathcal{I} \subseteq \mathcal{P}$ is called independent support of $\mathcal{P}$ if $\forall \sigma_1, \sigma_2 \in \satisfying{\varphi}, \sigma_1{_{\downarrow \mathcal{I}}} = \sigma_{2_{\downarrow \mathcal{I}}} \implies \sigma_{1_{\downarrow \mathcal{P}}} = \sigma_2{_{\downarrow \mathcal{P}}}$. In this paper, we focus on the design of efficient algorithmic techniques to compute $I$ for a given $\varphi$ and $\mathcal{P}$.

\subsection{ Applications: Counting and Sampling}
Given a Boolean formula $\varphi$ and a projection set $\mathcal{P} \subseteq X$, the problem of projected model counting seeks to compute the number of solutions of the formula $\exists Y \varphi$ where $Y = X \setminus \mathcal{P}$; alternatively, the solutions of the formula $\exists Y \varphi$ can be viewed as solutions of the formula projected on $\mathcal{P}$. Observe that projected model counting is a strict generalization of the classical problem of propositional model counting which focuses on the case when $\mathcal{P} = X$.  Similarly, the problem of projected sampling seeks to uniformly sample solutions of $\exists Y \varphi$.

Projected counting and sampling are fundamental problems in computer science with a wide variety of applications ranging from network reliability~\cite{DMPV17}, neural network verification~\cite{BSSMS19}, computational biology~\cite{SE19}, and software and hardware testing~\cite{NavEm05}. For example, given a neural network $\mathcal{N}$ and a property $\psi$, the problem of estimating how often the network satisfies the property $\psi$ reduces to projected counting~\cite{BSSMS19}.

From a theoretical viewpoint, projected counting is \#NP-complete; it is worth remarking that the problem of propositional model counting is \#P-complete~\cite{Valiant79}, and it is known that \#P $\subseteq $ \#NP.
 The hardness of \#P (and \#NP) motivated efforts towards approximation methods with $(\varepsilon,\delta)$-guarantees. The state of the art approximate techniques for counting and sampling are hashing-based~\cite{CMV13a,CMV13b,CMV14,SM19,SGM20}, seeking to combine the power of universal hash functions with advances in SAT solving.

The core idea of hashing-based counting techniques is to employ pairwise independent hash functions, also known as strongly universal hash functions, to partition the solution space into {\em roughly equal small} cells of solutions and then pick a cell randomly. The number of solutions in a cell can then be determined exactly by enumerating the solutions in the cell one by one, in case the cell is {\em small}. The number of solutions of the formula is then estimated as the number of solutions of a {\em randomly chosen small} cell multiplied by the number of cells. In case of sampling, we enumerate the solutions in the cell and chose one of the solutions uniformly at random.

To achieve estimates with rigorous $(\varepsilon,\delta)$-guarantees (formally defined in Section~\ref{sec:prelims}),  we use 3-wise independent hash functions\footnote{The hashing-based counting technique only require 2-wise independent hash functions but the current state of the art hashing-based sampling techniques rely on 3-wise independent hash functions}. The current state of the art techniques rely on XOR-based 3-wise independent hash functions. A randomly picked hash function $h: \{0,1\}^n \mapsto \{0,1\}^m$ is explicitly represented as a conjunction of $m$ randomly chosen XOR constraints, where each constraint is constructed by picking every variable with probability $\frac{1}{2}$. Therefore, the expected size of each XOR is $\frac{n}{2}$ where $n = |\mathcal{P}|$. Accordingly, the SAT solver is invoked to find solutions of a formula expressed as conjunction of the original formula, $\varphi$, and the XOR constraints.

Runtime performance of hashing-based counting and sampling techniques is primarily determined by the time taken by the SAT solver as over 99\% of the time is spent inside SAT calls. Accordingly, the past decade has witnessed a sustained effort in designing {\em sparse} XOR-based hash functions~\cite{GHSS07,EGSS14,CMV14,MA20}. An important advance in this direction was achieved by Chakraborty et al.~\cite{CMV14}, who proposed the notion of independent support and observed that one can construct XORs only over the given independent support. It is worth remarking that Chakraborty et al. had defined the notion of independent support $\mathcal{P} = X$ but one can see that the notion easily generalizes to arbitrary $\mathcal{P}$.

\subsection{Techniques to Identify Independent Support}
Ivrii et al.~\cite{IMMV15} showed that the problem of computation of Minimal Independent Support can be reduced to Group Minimal Unsatisfiable Subset (GMUS), and the corresponding tool, {\MIS}, was shown to scale to {\em moderately complex} instances.
Concurrently, Lagniez et al.~\cite{LM16,LM20} published a pre-processing technique, B+E, that combines the assumption-based framework offered by modern CDCL solvers with Padoa's theorem~\cite{P01} to identify an independent support. An important conceptual viewpoint put forth by Lagniez et al. was the argument of eschewing the search for minimal independent support and instead focus on efficiently finding an independent support that may not be necessarily minimal.

\subsection{Our Contributions}
Our primary contribution is the design of an efficient framework, called {\arjun}, to identify independent support. {\arjun} consists of two phases.
The first phase employs a syntactic gate recovery-based strategy while the second phase is based on the tight integration of the standard assumption-based CDCL framework with the invocations needed to execute on Padoa's Theorem~\cite{P01}.

 To demonstrate the efficacy of our approach, we perform a detailed empirical analysis over an extensive set of 1896  benchmarks arising from a diverse set of domains and employed in the analysis of counting and sampling techniques.
 We first showcase that with a timeout of 5000 seconds, {\arjun} can compute independent support for
 358 more instances than the state of the art tool, {\MIS}.

 We observe that {\ApproxMCFour} augmented with {\arjun} can count 1596 benchmarks, achieving an improvement of
 387 benchmarks over {\ApproxMCFour}. Similarly, we observe that {\UniGenThree} augmented with {\arjun} can sample 1354 benchmarks, achieving an improvement of
 319 benchmarks within the same time limit of 5000s. Furthermore, we observe a significant improvement in runtime performance as well, in particular, the PAR-2 scores\footnote{PAR-2 scores are used in the SAT competitions to measure performance. Each benchmark contributes a score that is the number of seconds used to finish (e.g. solve, count, sample) it, or in case of a timeout or memory out, twice the timeout in seconds. We then calculate the average score for all benchmarks, giving PAR-2.} for {\ApproxMCFour} and {\UniGenThree} augmented with {\arjun} are 1707 and 2957 while the PAR-2 scores for {\ApproxMCFour} and {\UniGenThree} are 3800 and  4717, respectively.

\paragraph{Organization} The rest of the paper is organized as follows: We provide a detailed background in Section~\ref{sec:prelims} and then present related work in Section~\ref{sec:relatedwork}. We then present a detailed algorithmic description of {\arjun} in Section~\ref{sec:algorithm}. We present an extensive empirical evaluation in Section~\ref{sec:evaluation} and finally conclude in Section~\ref{sec:conclusion}.
\section{Background}\label{sec:prelims}

Given a graph $G = (V,E)$, a feedback vertex set $S \subseteq V$ is the set of vertices whose removal makes the graph acyclic. For an edge $e=(u,v)$, we say the edge $e$ is outgoing from $u$ and incoming to $v$. For a directed graph $G$, we use $\mathsf{Root}(G)$ to denote the set of vertices in $G$ which do not have any incoming edge.

Let $X = \{x_1, x_2 \ldots x_n\}$ be the set of Boolean variables. Let a \emph{literal} be a Boolean variable or its negation. A formula, $\varphi$, defined over $X$ is known as a Conjunctive Normal Form (CNF), if $\varphi$ is a conjunction of \emph{clauses}, where each \emph{clause} is disjunction of literals.
 A \emph{satisfying assignment} $\sigma$ of $\varphi$ is a mapping of $X  \rightarrow \{0,1\}$, such that $\varphi$ evaluates True at $\sigma$. We often represent $\sigma$ as the set of literals.  We use $\sigma \models \varphi$ to denote $\sigma$ as a {\em solution} or a \emph{satisfying assignment} of $\varphi$. Furthermore, for $\mathcal{P} \subseteq X$, $\sigma_{\downarrow \mathcal{P}}$ represents the assignment of variables restricted to $\mathcal{P}$.    We denote the set of all witnesses of  $\varphi$ by $\satisfying{\varphi}$, and use $\ProjectSatisfying{\varphi}{\mathcal{P}}$ to indicate the projection of $\satisfying{\varphi}$ on $\mathcal{P}$.\\

\noindent {\bfseries Example:} consider $X = \{x_1, x_2, x_3\}$ and let $\sigma = (x_1, \neg x_2, x_3)$ (implying that $x_1$ and $x_3$ map to True while $x_2$ maps to False). Let $\mathcal{P} = \{x_1, x_2\}$, then $\sigma_{\downarrow P} = \{x_1, \neg x_2\}$.

\subsection*{Definability and Independent Support}

 \begin{definition}
 A subset of variables $\mathcal{I} \subseteq \mathcal{P}$ is an independent support of $\mathcal{P}$; if
 $\sigma_1, \sigma_2 \in \satisfying{\varphi}_{\downarrow P}, \text{ we have }  \sigma_{1_{\downarrow \mathcal{I}}} = \sigma_{2_{\downarrow \mathcal{I}}} \implies \sigma_{1_{\downarrow P}} = \sigma_{2_{\downarrow P}}$
 \end{definition}

Furthermore, $\mathcal{I} \subseteq \mathcal{P}$ is called minimal independent support of $\mathcal{P}$ if there does not exist $\hat{\mathcal{I}} \subset \mathcal{I}$ such that $\hat{\mathcal{I}}$  is an independent support.

Observe that if $\mathcal{I}$ is an independent support of $\mathcal{P}$, then we have $|\satisfying{\varphi}_{\downarrow \mathcal{I}}| = |\satisfying{\varphi}_{\downarrow \mathcal{P}}|$. The notion of independent support is related to the classical notion of definability. To this end, we first present the following two equivalent notions of definability.

\begin{definition}[Implicit Definability]\label{def:implicit}
	A variable $x \in \mathcal{P}$ is implicitly defined by $\mathcal{I}$ for the formula $\varphi$ if and only if $\forall \tau \in 2^\mathcal{I}$, we have $\varphi \wedge \tau \models x $ or $\varphi \wedge \tau \models \neg x$.
\end{definition}

\begin{definition}[Explicity Definability]
	A variable $x \in \mathcal{P}$ is explicitly defined by $\mathcal{I}$ for the formula $\varphi$ if and only if there exists $\phi(\mathcal{I})$ such that
	$\varphi   \models x \leftrightarrow \phi(\mathcal{I}) $
\end{definition}

\begin{lemma}[Beth's Theorem~\cite{B56}]
	A variable $x \in \mathcal{P}$ is explicitly defined by $\mathcal{I}$ for the formula $\varphi$ if and only if $x$ is implicitly defined by $\mathcal{I}$ for the formula $\varphi$.
\end{lemma}

 Since implicit and explicit definability are equivalent, we can omit referencing them and simply make statements such as: $x$ is defined by $\mathcal{I}$ for the formula $\varphi$. Furthermore, the following remark follows from the notion of Independent support.
 \begin{remark}
 	 If $\mathcal{I}$ is an independent support of $\mathcal{P}$ then all the variables  $\mathcal{P} \setminus \mathcal{I}$ are defined by $\mathcal{I}$.
 \end{remark}

\subsection*{Counting and Sampling}

 The problem of \emph{propositional model counting} is to compute
 $|\satisfying{\varphi}|$ for a given CNF formula $\varphi$.  A \emph{probably approximately correct}
 (or \PAC) counter is a probabilistic algorithm ${\ApproxCount}(\cdot,
 \cdot,\cdot)$ that takes as inputs a formula $\varphi$,  a tolerance $\varepsilon>0$, and a confidence $1-\delta \in
 (0, 1]$, and returns a count $c$ with $(\varepsilon,\delta)$-guarantees, i.e.,
 $\prob\Big[\frac{|\satisfying{\varphi}|}{(1+\varepsilon)} \le c \le
 (1+\varepsilon)|\satisfying{\varphi}|\Big] \ge 1-\delta$. Projected model counting is defined
 analogously using $\ProjectSatisfying{\varphi}{\mathcal{P}}$ instead of
 $\satisfying{\varphi}$, for a given projection
 set\footnote{Projection set has been referred to as sampling set in prior work~\cite{CMV14,SM20}} $\mathcal{P} \subseteq X$.

 A {\em uniform sampler} outputs a solution $y \in \satisfying{\varphi}$ such that $\prob[y \text{ is output}] = \frac{1}{|\satisfying{\varphi}|}$. An {\em almost-uniform sampler} relaxes the guarantee of uniformity and in particular, ensures that $\frac{1}{(1+\varepsilon)|\satisfying{\varphi}|} \leq \prob[y \text{ is output}] \leq \frac{1+\varepsilon}{|\satisfying{\varphi}|}$.

\section{Related Work}\label{sec:relatedwork}

Padoa's theorem provides necessary and sufficient condition to determine whether $x$ is defined by $\mathcal{I}$ for the formula $\varphi$. Let $\varphi(X)$ be defined on $X = \{x_1, x_2, \ldots x_n\}$, and $\mathcal{P}$ be of size $t$. Without loss of generality,  let $\mathcal{P} = \{x_1, x_2, \ldots x_t\}$.  We create another set of {\em fresh} variables $\hat{\mathcal{P}} = \{y_1, y_2, \ldots y_t\}$. Let $\varphi(\mathcal{P} \mapsto \hat{\mathcal{P}})$ represent the formula where every $x_i \in \mathcal{P}$ in $\varphi$ is replaced by $y_i \in \hat{\mathcal{P}}$.

\begin{lemma}[Padoa's Theorem~\cite{P01}]

	\begin{align*}
\psi(X,\hat{\mathcal{P}},i) := & \varphi(X) \wedge
\varphi(\mathcal{P} \mapsto \hat{\mathcal{P}}) \wedge
\bigwedge_{j=1\atop {j\neq i}}^{t} (x_j \leftrightarrow y_j) \wedge
x_i \wedge
\neg y_i
	\end{align*}
	A variable $x_i \in \mathcal{P}$ is  defined by $\mathcal{I}$ for the formula $\varphi$ iff $\psi(X,\hat{\mathcal{P}},i)$ is unsatisfiable.
\end{lemma}

\begin{remark}
For a given formula $\varphi$, if variable $x_i \in \mathcal{P}$ is  defined by $\mathcal{I}$  and a variable $x_j \in \mathcal{I}$ is defined by $\mathcal{I} \setminus x_j$, then $x_i$ is defined by $\mathcal{I} \setminus x_j$.
\end{remark}

Combining the above observation with Padoa's theorem, Lagniez et al.~\cite{LM16} proposed an iterative procedure that performs $|\mathcal{P}|$ calls to a SAT oracle to determine a minimal independent support. To improve the efficiency, the proposes to invoke a SAT solver with a set number of conflicts and to treat SAT or timeout as equivalent. Note that the presence of a pre-defined limit on conflicts causes the loss of guarantee of minimality for the independent support that is returned by the technique. However, such a loss of minimality is at the gain of efficiency in identifying a {\em small enough} independent support. Based on their paper, \cite{LM16} published the tool {\BplusE} that performs the extraction of minimal independent support given an input CNF.

In another line of work, Ivrii et al.~\cite{IMMV15} reduced the problem of minimal independent support to Group Minimal Unsatisfiable Subset (GMUS)~\cite{LS08}. While the past decade has witnessed significant advances in the development of efficient GMUS tools,
scalability remains a challenge. As a result, while Ivrii et al's proposed tool, {\MIS}, workings exceedingly well for easy- to moderate-complexity formulas, it has difficulties with harder CNF formulas.%

Recently, Slivovsky~\cite{S20} observed that the resolution proof for unsatisfiability of $\psi(X,\hat{\mathcal{P}},i)$ can be employed to generate the definition $\phi_i$ and $x_i$ such that $\varphi   \models x_i \leftrightarrow \phi_i(\mathcal{I}) $. Furthermore, Slivovsky used all such extracted $\phi_i$-s to perform pre-processing in the context of QBF.

\section{Algorithm}\label{sec:algorithm}
In this section, we delve into the primary technical contribution of this paper: {\arjun}, an efficient technique to compute independent support. {\arjun} consists of two phases. Each phase takes the formula $\varphi$ and a projection set $\mathcal{P}$ as input and returns a set $\mathcal{I}$ such that $\mathcal{I} \subseteq \mathcal{P}$ is an independent support of $\mathcal{P}$. The two phases can be composed sequentially, by feeding the output of one phase as an input projection set to another phase

The first phase employs a gate identification-based strategy while the second phase is based on a tight integration of assumption-based framework to efficiently perform invocations based on Padoa's Theorem. It is worth emphasizing that while Beth's theorem asserts that both implicit and explicit definability are equivalent notions, our framework seeks to exploit the observation that there exists two classes of definable (or dependent) variables: one for which it is easy to extract their explicit definitions while for the other class of variables, we rely on the check for implicit definability via Padoa's Theorem.

\subsection{Explicit Definability-based Identification}

Given a formula $\varphi$ and a projection set $\mathcal{P}$, we focus on finding the subset of variables $\mathcal{I} \subseteq \mathcal{P}$ along with the set of definitions $\Phi$ such that for every $x \in \mathcal{I}$, there is a corresponding definition $\phi \in \Phi$ such that $\phi$ is defined over $\mathcal{I}$. A possible method for the same can be to rely on Slivovky's observation of extractions of definitions from the resolution proofs but such a technique is computationally expensive given the requirement of SAT calls.

We observe that syntax-based gate identification techniques such as \cite{DBLP:conf/cp/OstrowskiGMS02} can be employed on the CNF to efficiently recover an incomplete set of definitions. Given the set of such definitions, we can extract $\mathcal{I}$ and $\Phi$ with the desired properties. Our explicit definability-based identification phase consists of three steps:
\begin{algorithm}
	\caption{$\explicit(\varphi,\mathcal{P})$}
\begin{description}
	\item[Step 1] Build a list of gate-definitions wherein for every gate $g$ is represented as a tuple of the form $(\ell, \mathsf{op, litList})$ such that $\mathsf{\ell = op(litList)}$, i.e., literal $\ell$ can be expressed as output of the gate corresponding to the operator $\mathsf{op}$ over the literal list $\mathsf{litList}$. \\

	\item[Step 2] Construct a directed graph $G = (V,E)$ where there is a vertex $v \in V$ for every variable $x \in \mathcal{P}$,  and for every gate $(\ell, \mathsf{op, litList})$, we have an edge from variables in $\mathsf{litList}$ to the variable of $\ell$. It is worth emphasizing that the gates are defined over $\mathsf{litList}$.\\

	\item[Step 3] Compute feedback vertex set $W$ of $G$ in a greedy fashion and return $\mathcal{I} = W \cup \mathsf{Root}(G)$ wherein $\mathsf{Root}(G)$ corresponds to the set of vertices in $G$ without any incoming edges.
\end{description}
\end{algorithm}
We now discuss {\bfseries Step 1} and {\bfseries Step 3} in more detail. First of all, we focus on two set of operators: AND $(\wedge)$ and XOR $(\oplus)$. We detect AND gates of length two while we detect XOR gates of length up to five, both over literals rather than variables.
Our advanced AND gates detection works up to length two only, since 2-long AND gates account for the overwhelming set of AND gates present in the benchmarks of interest. Since we focus on variables rather than literals, detection of OR gates is not required, since an OR gate is equivalent to an AND gate on the opposite literals due to De Morgan's laws~\cite{de-morgan-algebra}. For the algorithm used to efficiently detect AND gates, see Appendix~\ref{appendix:and-gate-recover}.

In case of XOR gates, we rely on the bloom-filter based XOR recovery algorithm proposed by~\cite{SM19}. We instantiate this recovery algorithm to find XORs of length up to five in order to limit potential explosion of search space. Unlike AND gates, the identification of XOR equation does not put restrictions on the outputs and inputs. In particular, observe that an XOR $\ell_1 \oplus \ell_2 \oplus \ell_3 = 1$ can be rewritten in the following ways: (1)
$\neg \ell_1 =  \ell_2 \oplus \ell_3$, (2) $\neg \ell_2 =  \ell_1 \oplus \ell_3$, and (3) $\neg \ell_3 =  \ell_1 \oplus \ell_2$. Therefore, given an XOR of length $k$, we extract $k$ definitions.

\begin{algorithm}[htb]
    \caption{$\mathsf{GreedyIndSearch(gates,}\mathcal{P}\mathsf{)}$}
    \label{alg:greedy}
    \begin{algorithmic}[1]
		\State $\mathsf{SortInc(\mathcal{P}, incidence)}$ \Comment{Most likely dependent first}

        \For {$u \in \mathcal{P}$} \Comment{Take most likely dependent variable}
        		\For {$g \in \mathsf{gates[u]}$}
	        		\State OK $\leftarrow$ True
	        		\For {$v \in g.\mathsf{literals}$}
	        			\If {$v \notin \mathcal{P}$}
	        				\Comment{Could lead to cycle, skip}
	        				\State OK $\leftarrow$ False
	        				\Break
	        			\EndIf
	        		\EndFor
	        		\If {OK}
	        		\State $\mathcal{P} \leftarrow \mathcal{P} \setminus u$
	        		\Break
	        		\EndIf
        	\EndFor
        \EndFor
    \end{algorithmic}
\end{algorithm}

We now return to {\bfseries Step 3}, performing a greedy search to compute $W \cup \mathsf{Root}(G)$ wherein W is a feedback vertex set of $G$. While ideally, we would like $W$ to be a minimal feedback vertex set, we trade off the minimality for runtime performance. The key strategy is to sort the vertices according to the number of edges incident onto them; i.e., we observe that the higher the number of edges incident upon a vertex $u$, the higher likelihood of $u$ to belong to $\mathcal{I}$.  We present the pseudocode of the greedy search in  Algorithm~\ref{alg:greedy}. The algorithm  $\mathsf{GreedyIndSearch}$ relies on the array $\mathsf{incidence}$ that is computed for each variable $u$ as the number of clauses containing $u$ or $\neg u$ in $\varphi$.

\subsection{Implicit Definability-based Identification}

As noted in Section~\ref{sec:prelims}, Lagniez et al. observed that one can iteratively identify independent support. We first extend  Lagniez et al.'s proposal to handle projection set and present the resulting pseudocode, called {\simpleSearch}, in Algorithm~\ref{alg:simplesearch}.
\begin{algorithm}[htb]
	\caption{$\simpleSearch(\varphi, \mathcal{P})$}\label{alg:simplesearch}
	\begin{algorithmic}[1]
		\State $Y \gets \CreateNewVars(X)$;
		$Z \gets \CreateNewVars(\mathcal{P})$
		\State $\psi \gets \varphi(X) \wedge \varphi(X \mapsto Y) \wedge \bigwedge_{i=1}^{|\mathcal{P}|} (z_i \rightarrow (x_i = y_i))$
		\State $\mathsf{solver.addConstraint(\psi)}$
		\State $\unknown \leftarrow \{1, 2, \ldots |\mathcal{P}| \} $;
		$\indep \leftarrow \emptyset$

		\Statex  \Comment{Sort most likely dependent last}
		\State $\mathsf{sortDesc(\unknown, incidence)}$
		\While{$\unknown \ne \emptyset$}
			\State \assumps.clear() \label{line:simple-setup-begin}
			\Statex \Comment{Take most likely dependent variable}
			\State $\mathsf{index} \leftarrow \unknown$.pop()
			\For{j $\in \indep$}
			\assumps.append($z_j$)
			\EndFor
			\For{j $\in \unknown$}
			\assumps.append($z_j$)
			\EndFor
			\State \assumps.append($x_{\mathsf{index}}$)
			\State \assumps.append($\neg y_{\mathsf{index}}$)\label{line:simple-setup-end}
			\State $\mathsf{ret \leftarrow solver}.\solve(\assumps)$\label{line:simple-solve}
			\If {$\mathsf{ret} \not =$ UNSAT}\label{line:simple-check}
			\indep.append(${\mathsf{index}}$)
			\EndIf
		\EndWhile
		\State \Return \indep
	\end{algorithmic}
\end{algorithm}

\begin{algorithm}[h!]
	\caption{$\integratedImplicit(\varphi,P)$ }
	\label{alg:integratedsolve}
	\begin{algorithmic}[1]
		\State $Y \gets \CreateNewVars(X)$; $Z \gets \CreateNewVars(\mathcal{P})$ \label{line:integrate-setup-begin}
		\State $\psi \gets \varphi(X) \wedge \varphi(X \mapsto Y) \wedge \bigwedge_{i=1}^{|P|} (z_i \rightarrow (x_i = y_i))$
		\State $\mathsf{solver.addConstraint(\psi)}$

		\State $\unknown \leftarrow \{1, 2, \ldots |P| \} $;  $\indep \leftarrow \emptyset$
		\State $\mathsf{sortDesc(\unknown, incidence)}$
		\For{$i \in \unknown$} \assumps.push($z_i$)
		\EndFor
		\label{line:integrate-setup-end}
		\While{True}
		\State \unskip \texttt{start:}
		\State branch $\leftarrow$ None
		\For{i $\leftarrow$ branch\_depth; i $<$ assumps.size() AND branch is None; i++}
		\State lit $\leftarrow$ \assumps[i]; $\mathsf{index} \gets i+1$
		\If {value(lit) is False} \label{line:integrated-unsat-begin}

		\State \assumps.pop(); \assumps.pop()
		\If {\assumps.size() = \indep.size()} \Return \indep
		\EndIf

		\State \assumps.push($x_{\mathsf{index}}$)
		\State \assumps.push($\neg y_{\mathsf{index}}$)

		\Continue \label{line:integrated-unsat-end}
		\EndIf
		\If {value(lit) is True}
		\State new\_decision\_level() %
		\Continue
		\EndIf
		\State branch $\leftarrow$ lit
		\EndFor
		\If{branch is None}

		\State branch $\leftarrow$ pick\_branch()
		\If {branch is None OR conflict\_limit()} \label{line:integrated-sat-begin}
		\State \assumps.pop();  \assumps.pop()
		\State \indep.append($x_{\mathsf{index}}$)
		\State splice $\leftarrow$ \indep.size()
		\State assumps.insert(splice, $z_{\mathsf{index}}$)
		\State backtrack\_until(splice)
		\If {\assumps.size() = \indep.size()} \Return \indep
		\EndIf

		\State \assumps.push($x_{\mathsf{index}}$);
		\State \assumps.push($\neg y_{\mathsf{index}}$)

		\State \textbf{go to} \texttt{start} \label{line:integrated-sat-end}
		\EndIf
		\EndIf

		\State new\_decision\_level()
		\State enqueue(branch)
		\State \unskip \texttt{prop:}
		\State ret $\leftarrow$ propagate() \label{line:integrated-prop-begin}
		\If {ret = conflict}
		\State analyze\_conflict();
		\State backtrack\_until(backjmp\_level)

		\State \textbf{go to} \texttt{prop}
		\EndIf
		\If {should\_restart()}
		\State backtrack\_until(assumps.size()) \label{line:integrated-prop-end}
		\EndIf
		\EndWhile
	\end{algorithmic}
\end{algorithm}

{\simpleSearch} takes in a formula $\varphi$ and a projection set $\mathcal{P}$, and returns the independent support $\mathcal{I}$. Without loss of generality, we assume that $\mathcal{P}= \{x_1, x_2, \ldots x_{|P|}\}$ (note that simple variable renaming suffices to achieve such a $\mathcal{P}$). The key idea is to construct $\psi$ and perform iterative {\solve} queries over $\psi$. The standard method is to use assumption-based framework where the solver is required to solve the formula under the set of assumptions expressed as assignment to variables.  In {\simpleSearch}, we maintain two sets: ${\unknown}$, the set of variables that are yet to be classified as dependent or independent and  $\mathcal{I}$, the set of variables that we have classified as belonging to the independent support. As will be observed later, we would ideally like to sort the variables in such a way that the variables belonging to the independent support are queried at the very end.

Note that variable $x_{\mathsf{index}}$ can be defined in terms of $\unknown \cup \indep$ if and only if $\psi \wedge x_{\mathsf{index}} \wedge \neg y_{\mathsf{index}}$ is UNSAT under the assumption of setting all $z_i \in  \unknown \cup \indep$ to True. Since some SAT calls may be very expensive, instead of invoking the solver to completion, we set a cutoff on the number of conflicts, and therefore, the solver may return SAT, UNSAT, or timeout. To account for timeout, we check, in line~\ref{line:simple-check}, whether ret $\not = $ UNSAT instead of checking ret $=$ SAT.

We now seek to understand the key bottleneck for scalability of {\simpleSearch}:  the call to SAT solver on line~\ref{line:simple-solve}. To this end, we first seek to understand how modern CDCL-based SAT solvers implement the {\solve} procedure.
Observe that every invocation of $\solve$ on line~\ref{line:simple-solve} of {\simpleSearch} would require insertion of the set of assumptions of the size $|\mathcal{I} \cup \unknown|$ and the resulting propagations. Since we invoke {\solve} $|\mathcal{P}|$ times, the underlying solver must deal with $\frac{|\mathcal{P}|^2}{2}$ insertions and the corresponding propagations. To put this into perspective, if $|\mathcal{P}| = 50000$, then we have over a billion calls to propagate, a relatively expensive operation.  At this point, observe that the variable appearing first in ${\unknown}$ is inserted and propagated for all except one $\solve$ call. Therefore, we focus on addressing the performance bottleneck via eliminating redundant work. To this end, the key idea is to pursue a tight integration of {\simpleSearch} and the {\solve} algorithm, combining the two into a single algorithm.
For the interested reader, {\solve}, as per the seminal MiniSat paper~\cite{ES03} is explained in Appendix~\ref{appendix:cdcl-algo}.

The pseudocode for the integrated approach, called {\integratedImplicit}, is presented in Algorithm~\ref{alg:integratedsolve}. We assume that $\varphi$ is satisfiable, else $\mathcal{I} = \emptyset$ can be returned. Similar to {\simpleSearch}, we construct the formula $\psi$ based on the input formula $\varphi$. The high-level structure of {\integratedImplicit} is similar to {\simpleSearch} with crucial difference arising in the low-level technical details of how to handle the cases when UNSAT, SAT, or the timeout limit is reached.

We now focus on the UNSAT case, i.e., when $x_{\mathsf{index}}$ is shown to be dependent (lines~\ref{line:integrated-unsat-begin}--~\ref{line:integrated-unsat-end}).  The key observation is that whenever $\varphi$ is satisfiable, then  $\psi \wedge \bigwedge_{i=1}^{\mathsf{index}} z_i$ is satisfiable. (Observe that $\{z_i\}_{i=1}^{\mathsf{index}}$ is conjuncted via {\assumps}). Therefore, if $\psi \wedge \bigwedge_{i=1}^{\mathsf{index}} z_i \wedge x_{\mathsf{index}} \wedge \neg y_{\mathsf{index}} $ is unsatisfiable, then we need to only remove the three assumptions, namely $\{x_{\mathsf{index}}, \neg y_{\mathsf{index}}, z_{\mathsf{index}}\}$ and do not need to backtrack to decision level zero.

In case of SAT (i.e., no more variables to branch on) or when the conflict limit is reached, we want to insert $z_{\mathsf{index}}$ into our {\assumps} such that it is never popped during the rest of the execution. To this end, we insert  $z_{\mathsf{index}}$ at the index determined by the current size of $\mathcal{I}$ and backtrack there, then continue running.

To summarize, {\integratedImplicit} effectively avoids backtracking more than 3 levels except in case of (1) a regular conflict (2) restarting, or (3) an independent variable is found. In case of a restart, SAT solvers normally go back to decision level 0 but here, that would deterministically re-create what has already been decided and propagated until decision level $\assumps$.size(), so we go back there instead. The usage of sorting of variables based on incidence, defined as the number of clauses containing the variable or its negation in $\varphi$ as per \cite{LM16} ensures that in practice, dependent variables (which are typically the vast majority of variables) are popped first. Therefore, in practice, we achieve a reduction from quadratic to linear in the number of propagation calls.

\subsection{{\arjun}: Putting It All Together}
Our proposed technique, {\arjun}, consists of combining  the two phases {\explicit} and {\integratedImplicit}, sequentially. As noted earlier, both {\greedyIndSearch} and {\integratedImplicit} take a formula $\varphi$ and a projection set $\mathcal{P}$ and return $(\varphi,\mathcal{I})$, where $\mathcal{I}$ is the independent support of $\mathcal{P}$. Observe that  if $\mathcal{I}_1$ is independent support of $\mathcal{P}$ and $\mathcal{I}_2$ is independent support of $\mathcal{I}_1$, then $\mathcal{I}_2$ is independent support of $\mathcal{P}$. Therefore, the two phases can be combined in an arbitrary order as the output of one phase can be fed as the projection set for another phase. Our implementation of {\arjun} invokes {\explicit} before {\integratedImplicit}.

\section{Empirical Evaluation}\label{sec:evaluation}
We developed a prototype implementation of {\arjun}\footnote{The tool is available open-source at \url{https://github.com/meelgroup/arjun}}.
The experiments were conducted on a high performance computer cluster, each node consisting of 2xE5-2690v3 CPUs with 2x12 real cores and 96GB of RAM, i.e 4GB limit per run. Note that this memory limit played no significant role, hence we will no longer reference it.

To evaluate the performance and the quality of independent support computed by {\arjun}, we conducted a comprehensive study on the state of the art counter
{\ApproxMCFour}  and sampler {\UniGenThree}. {\ApproxMCFour} is a highly competitive model counter, a version of which won the 2020 model counting competition in the projected counting track. The 2021 competition sought to focus on exact techniques and consequently, changed $\varepsilon = 0.1$ to $\varepsilon = 0.01$. Even then, {\ApproxMCFour}-based entry achieved 3rd place. As described during the competitive event of SAT 2021, had $\varepsilon$ been set to $0.05$, the {\ApproxMCFour}-based entry would have won the competition. All prior applications and benchmarking for approximation techniques have been presented with $\varepsilon=0.8$ in the literature.

For our evaluation, we used 1896 benchmarks as released by Soos and Meel~\cite{SM20}. It  comprises of a wide range of application areas including probabilistic reasoning, plan recognition, DQMR networks, ISCAS89 combinatorial circuits, quantified information flow, program synthesis, functional synthesis, logistics, and the like. The past few years have witnessed a surge of interest in projected counting and sampling; accordingly, 801 out of 1896 benchmarks specify a projection set. These benchmarks are tailored to counting and sampling, and are satisfiable. However, {\arjun} correctly handles UNSAT instances, returning an empty independent support, if it does not time out.

The prior state of the art approach, {\BplusE}, computes independent support only for the case when $\mathcal{P}=X$.
On the other hand, {\MIS} by~\cite{IMMV15} can compute independent support for an arbitrary $\mathcal{P}$ but is often significantly slower than {\BplusE} for the case when $\mathcal{P}=X$. Therefore, to ensure a comprehensive comparison, we experiment with both {\BplusE} and {\MIS}. When we perform empirical evaluation of {\arjun} vis-a-vis {\BplusE}, we ignore the $\mathcal{P}$ supplied with the instance and instead set $\mathcal{P}=X$. In case of empirical evaluation with {\MIS}, we use the $\mathcal{P}$ as supplied by the instances.

To understand the impact of independent support computation on ApproxMC, we can not ignore $\mathcal{P}$ since under standard complexity theoretic assumptions, transformation to an equivalent instance without projection set (i.e., setting $\mathcal{P}=X$) would entail an exponential blow-up as such a transformation amounts to quantifier elimination. Therefore, our empirical evaluation performs comparisons of {\ApproxMCFour}  and {\UniGenThree}\footnote{Latest recommended versions of {\ApproxMC} and {\UniGen}} preceded by {\arjun} and {\MIS} respectively, owing to {\BplusE}'s lack of support for projection. To put our performance improvements in perspective, we also evaluated the performance of {\ApproxMCThree} and {\UniGenTwo} on our benchmarks. In line with prior studies, we set $\varepsilon = 0.8$ and $\delta = 0.2$ for all the versions of {\ApproxMC}; in case of (all the versions of) {\UniGen}, we set $\varepsilon = 16$.

\subsubsection*{Research Questions}

Our empirical study sought to answer the following research questions: \textbf{RQ 1.} How does the runtime performance and the size of independent supports computed by {\arjun} compare via-a-vis to prior state of the approaches? \textbf{RQ 2.} How do different phases affect the runtime performance of {\arjun}? \textbf{RQ 3.} How does the augmentation of {\arjun} affect the runtime performance of hashing-based counting and sampling tools?

\subsubsection*{Summary of Results}

Overall, we observe that {\arjun} significantly outperforms {\BplusE} and {\MIS} in runtime performance. Furthermore, while we observe the critical importance of both phases {\explicit} and {\integratedImplicit}, the empirical analysis shows that the incremental impact of {\integratedImplicit} is higher than that of {\explicit}. We then observe that  {\ApproxMCFour} augmented with {\arjun} can count 1596 benchmarks, achieving an improvement of 387 benchmarks over {\ApproxMCFour}. Similarly, we observe that {\UniGenThree} augmented with {\arjun} can sample 1354 benchmarks, achieving an improvement of 319 benchmarks within the same time limit of 5000s. Furthermore, we observe a significant improvement in runtime performance as well, in particular, the PAR-2 scores for {\ApproxMCFour} and {\UniGenThree} augmented with {\arjun} are 1707 and 2957 while the PAR-2 scores for {\ApproxMCFour} and {\UniGenThree} are 3800 and 4717, respectively.

\subsection{Comparison of {\arjun} with  {\MIS} and {\BplusE}}

\begin{figure}[tb]
	\centering
	\includegraphics[width=\columnwidth]{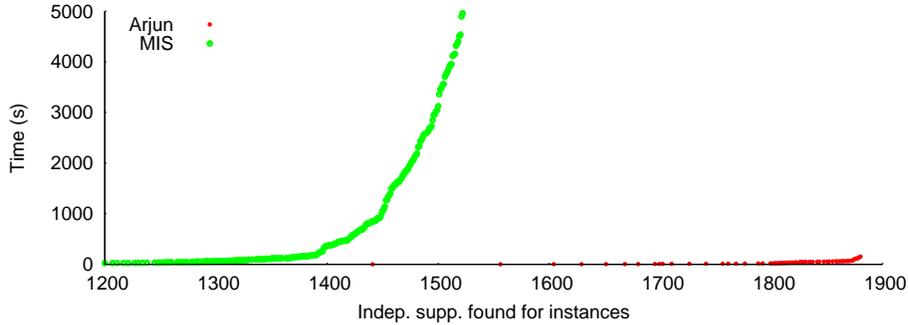}
	\caption{{\arjun} vs. {\MIS} independent support calculation times. %
	}
	\label{fig:mis-arjun-withindep-main}
\end{figure}

\begin{figure}[tb]
	\centering
	\includegraphics[width=\columnwidth]{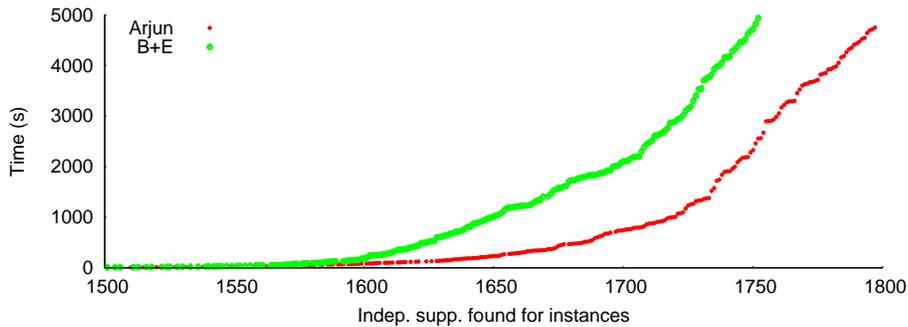}
	\caption{{\arjun} vs {\BplusE} independent support calculation times. Here, the projection set was discarded for an apples-to-apples comparison. Both systems used a 500 conflicts/variable timeout.}
	\label{fig:arjun-be}
\end{figure}

We present the runtime performance of {\arjun} vis-a-vis {\MIS} via cactus plot in Figure~\ref{fig:mis-arjun-withindep-main}. Observe that while {\arjun} could compute independent support for 1880 instances, {\MIS} could do the same for only 1522 instances within the same time limit. To illustrate the runtime performance difference, {\arjun} computed independent support for the same number of instances (i.e. 1522) in under 1.57s that {\MIS} took 5000s.

Even though from complexity theoretic viewpoint, we can not ignore $\mathcal{P}$,  we sought to further extensively analyze performance of {\arjun}.  To this end, we performed empirical evaluation vis-a-vis {\BplusE} by disregarding the user-provided projection set and instead setting $\mathcal{P}=X$ due to {\BplusE}'s inability to perform independent support computation for a non-trivial $\mathcal{P}$. We present the runtime performance cactus plot in Figure~\ref{fig:arjun-be}. While it took 2101 seconds for {\BplusE} to compute the independent support of 1700 instances, for {\arjun} it only took 741 seconds to do the same.

\begin{figure}[tb]
	\centering
	\subfloat{{
	\includegraphics[width=0.5\columnwidth]{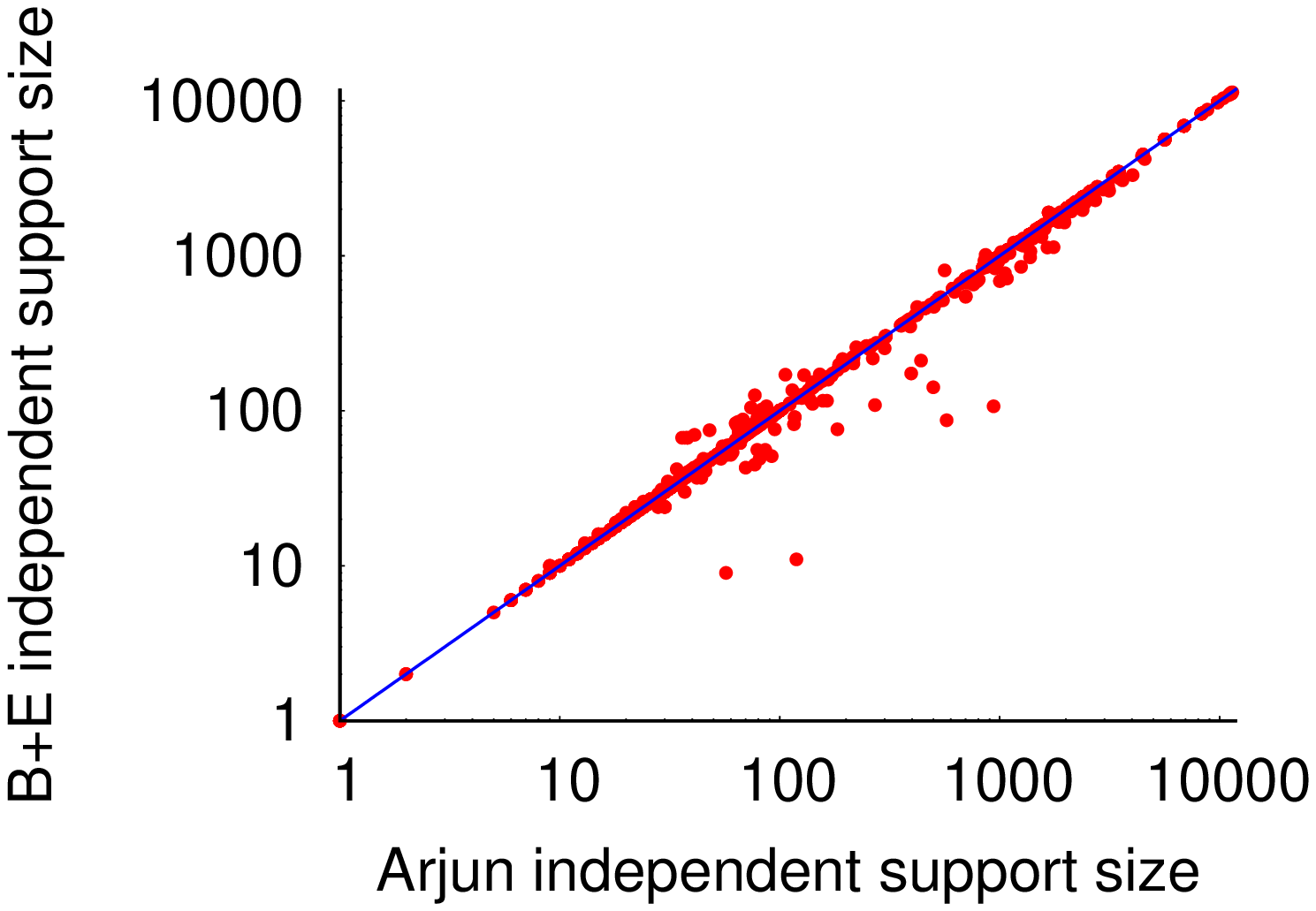}
	}}
	\subfloat{{
	\includegraphics[width=0.5\columnwidth]{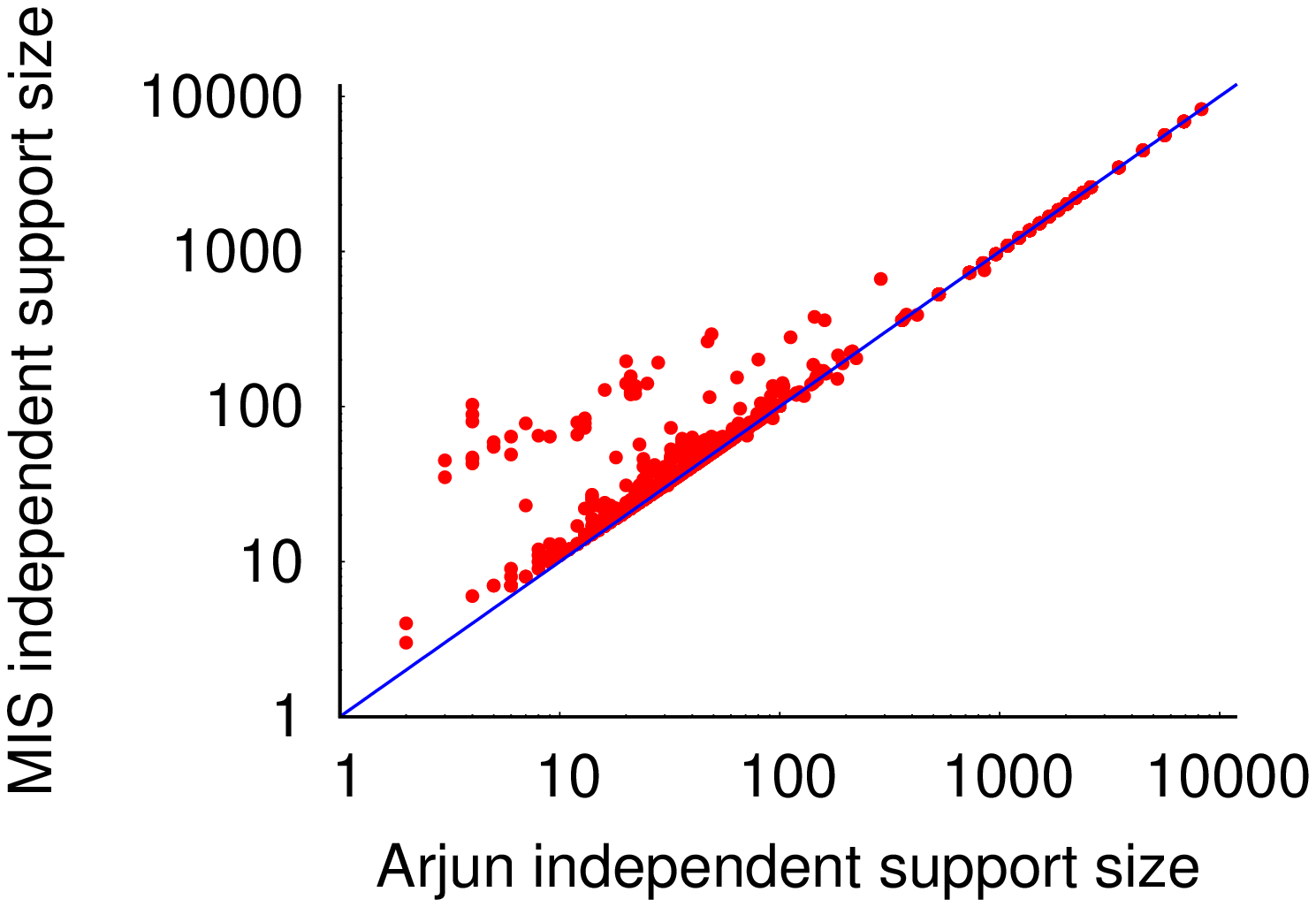}
	}}
	\caption{Computed independent support sizes of benchmarks
	}
	\label{fig:scatter-indeps}
\end{figure}

At this point, one may wonder about the size of independent supports computed by {\arjun}, {\MIS}, and and {\BplusE}. To this end, we present scatterplot of the size of independent support found by {\arjun}, {\MIS}, and {\BplusE} in Figure~\ref{fig:scatter-indeps}. Since {\arjun} found an independent support for a greater number of instances within the 5000s time limit, we only show the size of independent supports for instances that both {\arjun} and {\BplusE} (resp. {\arjun} and {\MIS}) finished on, for a true apples-to-apples comparison.

\subsection{Impact of {\explicit} and {\integratedImplicit}}
\begin{figure}[tb]
	\centering
	\includegraphics[width=\columnwidth]{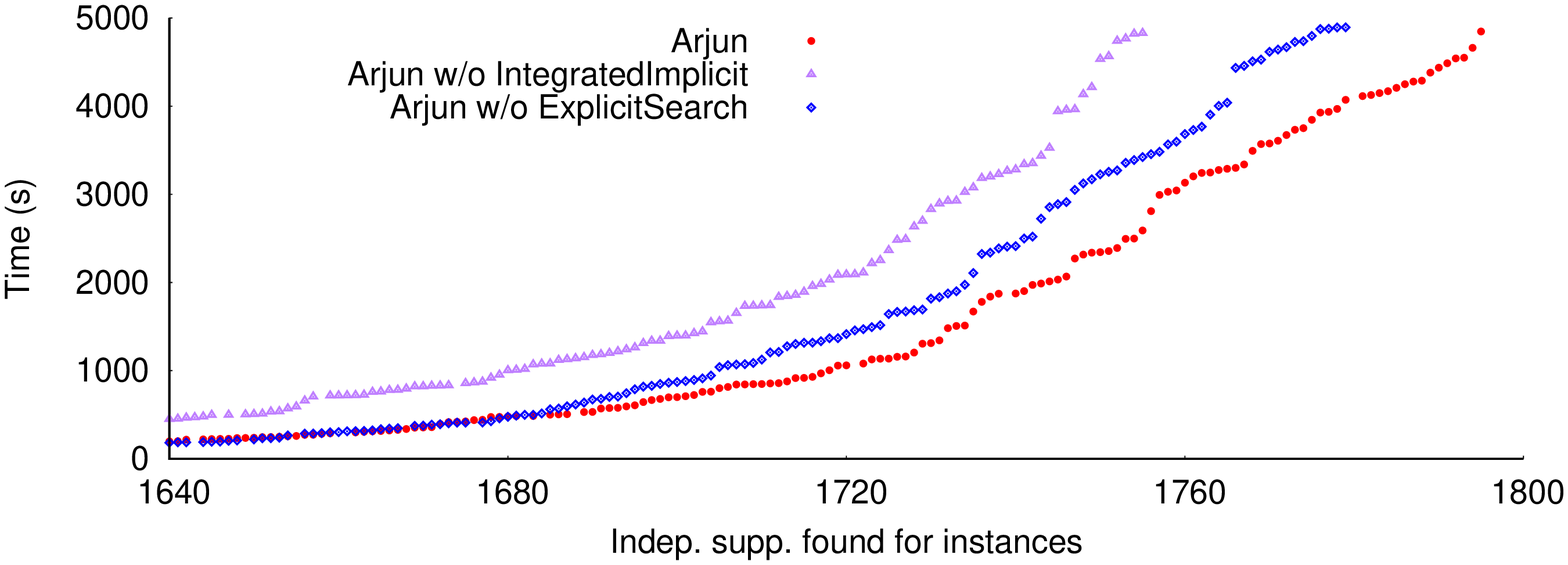}
	\caption{Impact of {\explicit} and {\integratedImplicit} on {\arjun}}
	\label{fig:arjun-phases}

\end{figure}

To understand the impact of {\explicit} and {\integratedImplicit}, we present the runtime performance of different versions of {\arjun} in Figure~\ref{fig:arjun-phases}. The curve ``{\arjun} w/o $\mathsf{IntegratedImplicit}$'' refers to the version of {\arjun} with $\mathsf{SimpleSearch}$ instead of {\integratedImplicit}, while ``{\arjun} w/o $\mathsf{ExplicitSearch}$'' refers to the version of {\arjun} without explicit definability-based identification. We make two observations: (1) both phases play a crucial role in the performance of the tool, and (2) the impact of {\integratedImplicit} is higher than that of {\explicit}.

\subsection{Impact of {\arjun} on {\ApproxMCFour} }
\begin{figure}[tb]
	\centering
	\includegraphics[width=\columnwidth]{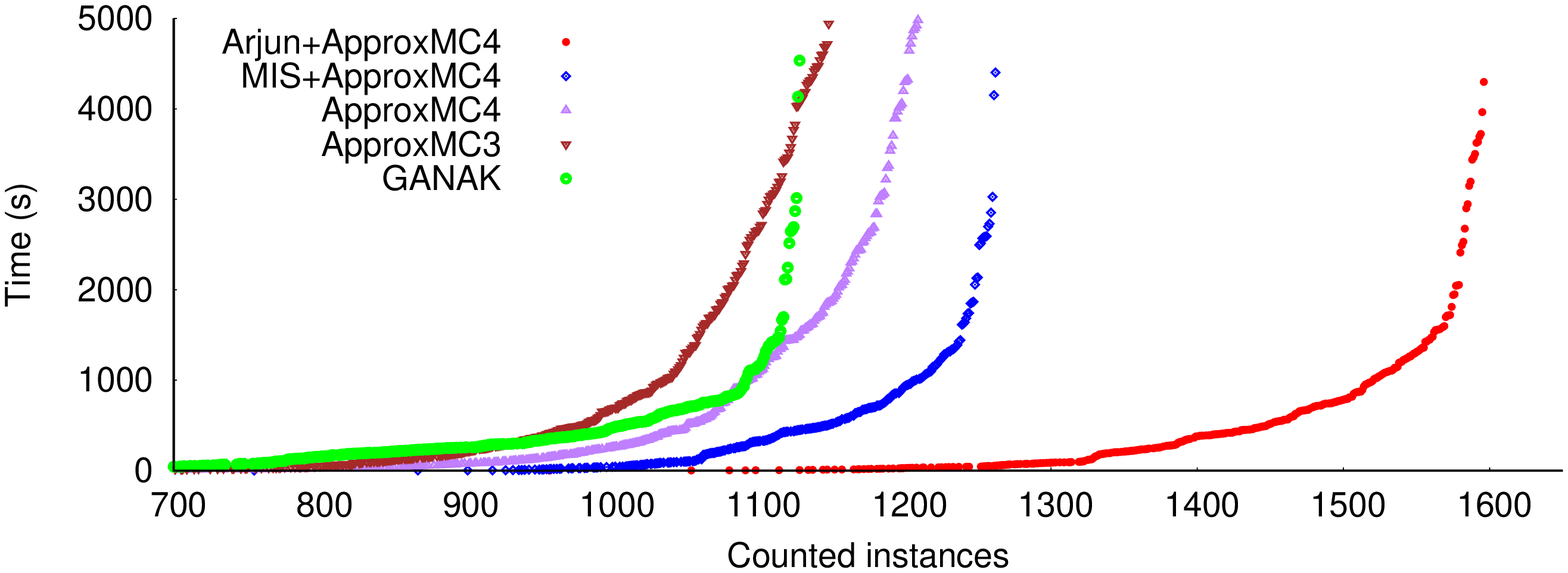}
	\caption{$\mathsf{GANAK}$ exact model counter~\cite{SRSM19}, {\ApproxMC}, {\arjun}+{\ApproxMC}, and {\MIS}+{\ApproxMC} projected model counting time%
	}
	\label{fig:appmc}
\end{figure}

Figure \ref{fig:appmc} shows the cactus plot of {\ApproxMCThree}, {\ApproxMCFour}, {\MIS}+{\ApproxMCFour}, and {\arjun}+{\ApproxMCFour} (cumulative time). Here, {\arjun} or ${\MIS}$ was first ran on the benchmark, and the resulting independent support was provided to {\ApproxMCFour}.
{\arjun}+{\ApproxMCFour} could count 1596 instances while {\ApproxMCFour} and {\MIS}+{\ApproxMCFour} could only count 1209 and 1262 instances, respectively within the same 5000s timeout. Overall, we observe an improvement of 334 instances. To put this into context, we also plot the curve corresponding to {\ApproxMCThree}, which demonstrates significant improvement {\arjun} provides relative to {\ApproxMCFour} vs {\ApproxMCThree}.
Solving speed improvement is substantial: with a timeout of only 34 seconds, {\arjun}+{\ApproxMCFour} could solve as many instances as {\ApproxMCFour} on its own under 5000 seconds. %

\begin{figure}[tb]
	\centering
	\includegraphics[width=\columnwidth]{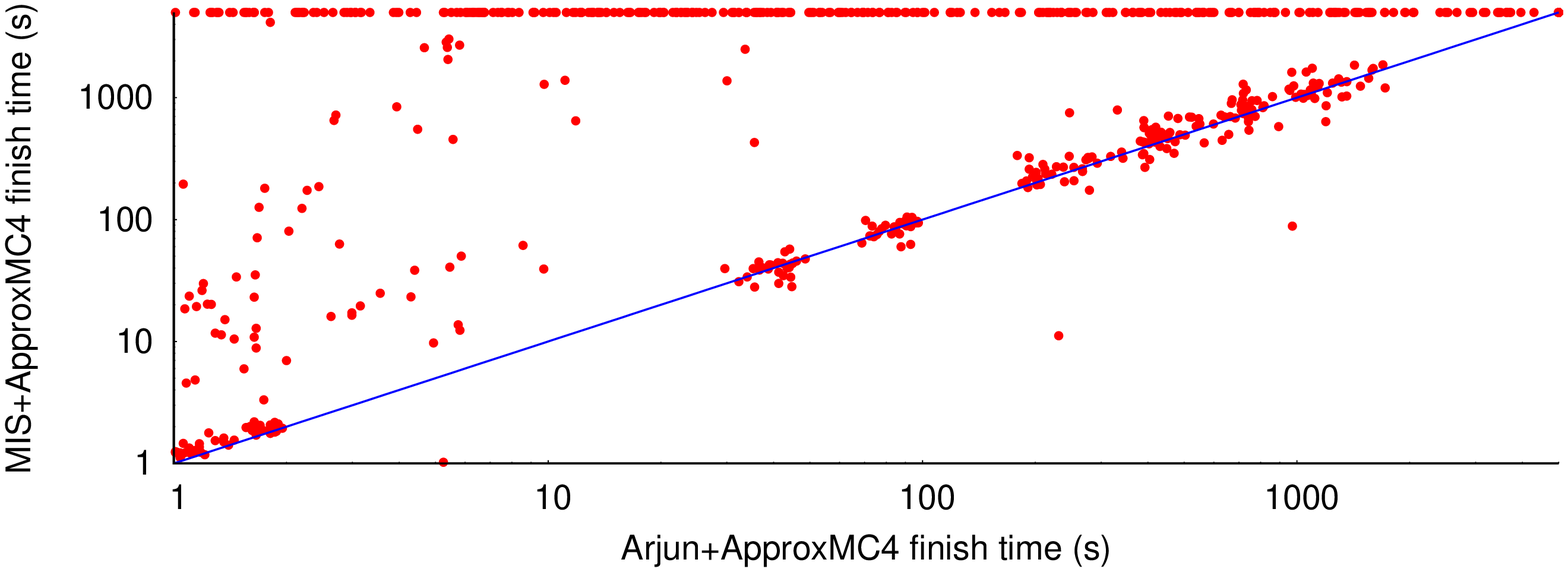}
	\caption{{\arjun}+{\ApproxMCFour} vs. {\MIS}+{\ApproxMCFour} time%
	}
	\label{fig:approxmc-arjun-mis}
\end{figure}

To further compare the runtime on a per instance basis, we present the scatter plot of {\arjun}+{\ApproxMCFour} vis-a-vis {\MIS}+{\ApproxMCFour} in Figure~\ref{fig:approxmc-arjun-mis}. We observe that on a per instance basis {\arjun}+{\ApproxMCFour} is almost always faster than {\MIS}+{\ApproxMCFour}.

This demonstrates the importance of efficient computational technique that can find small independent support. Recall that {\ApproxMCFour} adds XORs over the independent support in order to perform approximate counting. Hence, a smaller independent support leads to faster propagation and on average smaller conflict clauses. %

\subsection{Impact of {\arjun} on {\UniGenThree}}

\begin{figure}[tb]
	\centering
	\includegraphics[width=\columnwidth]{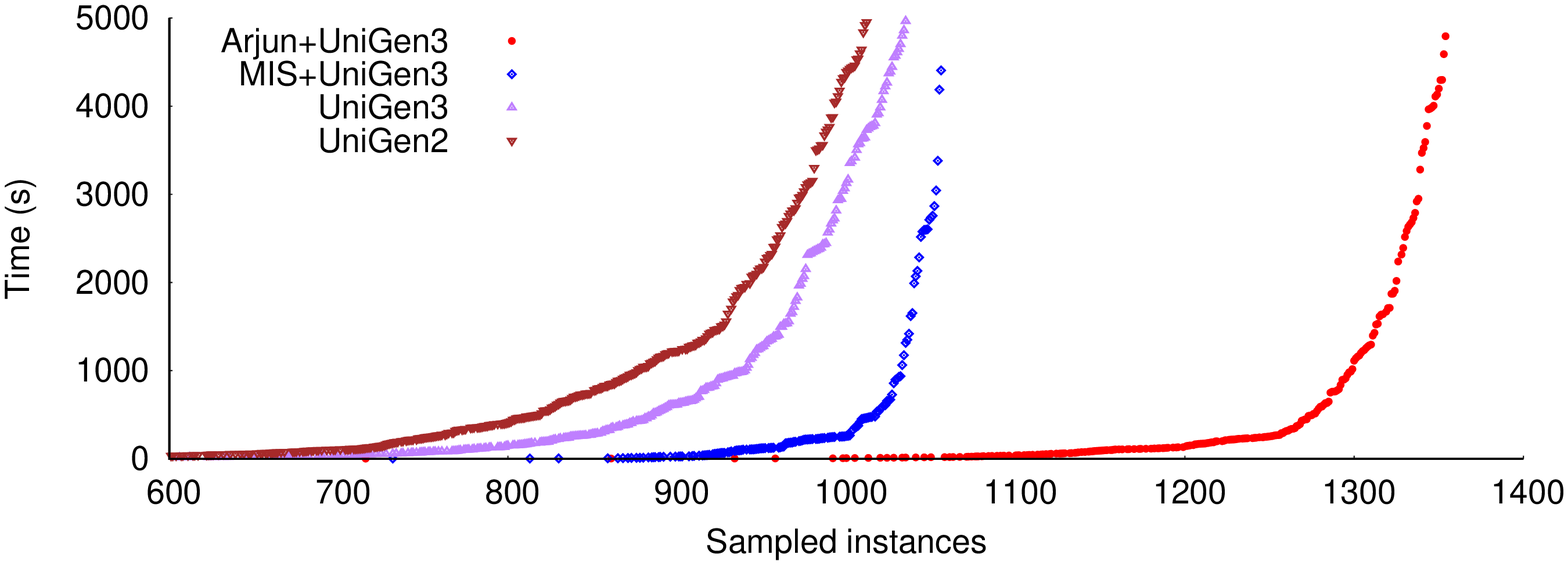}
	\caption{Runtime performance of {\UniGen}, {\MIS}+{\UniGen}, and {\arjun}+{\UniGen}}
	\label{fig:unigen}
\end{figure}

Figure \ref{fig:unigen} shows the cactus plot of {\UniGenTwo}, {\UniGenThree},  {\MIS}+{\UniGenThree}, and {\arjun}+{\UniGenThree} (cumulative time). Overall, {\arjun}+{\UniGenThree} samples 1354 instances, 298 more than {\MIS}+{\UniGenThree} within the same 5000s timeout. To put such a performance improvement into context, observe that the difference between {\UniGenTwo} and {\UniGenThree} in terms of number of sampled instances is significantly less than {\MIS}+{\UniGenThree} vs {\arjun}+{\UniGenThree}.
\section{Conclusion}\label{sec:conclusion}
In this paper, we focused on the problem of computation of independent support for a given formula owing to its importance for hashing-based counting and sampling techniques. Our algorithmic framework, {\arjun}, consists of two phases that seek to take advantage of implicit and explicit definability. The extensive empirical evaluation shows that {\arjun} achieves significant performance improvement over prior state of the art. Furthermore, augmenting  with {\arjun} the state of the art hashing-based counter {\ApproxMCFour} and sampler {\UniGenThree} leads to significant performance improvements.

Given the recent interest in developing counters for SMT, an interesting direction of future work would be to extend {\arjun} to handle SMT. Another interesting direction of research would be to integrate elimination techniques, due to Lagniez et al.~\cite{LM16}, over definable variables to simplify the given formula. Such instance simplification techniques have shown to produce significant runtime performance improvement in exact model counters by Sharma et al.~\cite{SRSM19}.
\section*{Acknowledgments}
This work was supported in part by National Research Foun-
dation Singapore under its NRF Fellowship Programme[NRF-
NRFFAI1-2019-0004 ] and AI Singapore Programme [AISG-
RP-2018-005], and NUS ODPRT Grant [R-252-000-685-13].
The computational work for this article was performed on
resources of the National Supercomputing Centre, Singapore
(https://www.nscc.sg).
\clearpage

\appendix

\section{Appendix}
\subsection{Fast Syntactic AND Gate Recovery From a CNF}
\label{appendix:and-gate-recover}
When trying to perform syntactic AND recovery from a CNF, we are looking for the following pattern in the CNF. First, there must exists a set of binary clauses
\begin{align*}
\mathsf{\neg outLit \vee inLit_1}, \mathsf{\neg outLit \vee inLit_2}, \ldots \mathsf{\neg outLit \vee inLit_n}
\end{align*}
and secondly, a longer clause
\begin{align*}
\mathsf{outLit \vee \neg inLit_1 \vee  \neg inLit_2 \vee \ldots \neg inLit_n}
\end{align*}
These together indicate the AND gate
\begin{align*}
\mathsf{outLit \leftrightarrow inLit_1 \wedge inLit_2 \wedge \ldots inLit_n}
\end{align*}

The currently best published work on syntactic gate recovery from a CNF is by \cite{gate-recovery-cnf}. It presents a flexible algorithm to discover a wide variety of gates. However, a faster algorithm is possible if only AND gates need to be discovered. To this end, we developed $\mathsf{find\_AND\_gates\_sweep}$ shown in Algorithm~\ref{alg:and-gate-search}.

Algorithm~\ref{alg:and-gate-search} works as follows. First, it clears out the return value $\mathsf{gates}$, then calls $\mathsf{sweep\_AND\_gates\_lit}$ for all possible output literals $\mathsf{outLit}$. In turn, $\mathsf{find\_gate\_out}$ in Algorithm~\ref{alg:find-gates-out} finds all possible AND gates that $\mathsf{outLit}$ can be an output of. To this end, it marks the literals that $\mathsf{\neg outLit}$ is in binary clauses with. If no literal has been marked, it is not possible to find any AND gates with $\mathsf{outLit}$ as an output in a purely syntactic way and the algorithm early aborts. Otherwise, it goes through all the clauses that $\mathsf{outLit}$ is part of, and ensures that all other literals in the clause have a marking on their opposite polarity. If it finds such a clause, it saves the clause  and literal pair as a new AND gate. Finally, it clears up the markings it has set.

\begin{algorithm}[htb]
	\caption{$\mathsf{find\_AND\_gates\_sweep(occlist, variables)}$}
	\label{alg:and-gate-search}
	\begin{algorithmic}[1]
		\State $\mathsf{gates}$ $\leftarrow$ 0
		\State $\mathsf{marker}$ $\leftarrow$ [0\ldots0]
		\For{$\mathsf{var \in variables}$}
		    \State $\mathsf{outLit} \leftarrow \mathsf{var}$
			\State $\mathsf{find\_gate\_out(outLit, gates, occlist)}$
			\State $\mathsf{find\_gate\_out(\neg outLit, gates, occlist)}$
		\EndFor
		\Return gates
	\end{algorithmic}
\end{algorithm}
\begin{algorithm}[htb]
 	\small
	\caption{$\mathsf{find\_gate\_out(outLit, gates, occlist, marker)}$}
	\label{alg:find-gates-out}
	\begin{algorithmic}[1]
	      \State $\mathsf{to\_clear \leftarrow} []$
		  \For{$\mathsf{occ \in occlist[\neg outLit]}$}
		  	\State $\mathsf{clause} \leftarrow \mathsf{get\_clause(occ)}$
		    \If{$\mathsf{len(clause) = 2}$}
		      \State $\mathsf{lit2} \leftarrow \mathsf{find\_other\_lit(clause, lit)}$
		      \State $\mathsf{marker[lit2] \leftarrow 1}$
		      \State $\mathsf{to\_clear.append(lit2)}$
		    \EndIf
		  \EndFor

		  \If {$\mathsf{to\_clear}$ = []}
		    \Return
		  \EndIf

		  \For{$\mathsf{occ \in occlist[outLit]}$}
            \State $\mathsf{cl \leftarrow get\_clause(occ)}$
		    \If{$\mathsf{len(clause) > 2}$}
		      \State $\mathsf{ok \leftarrow}$ True
		      \For{$\mathsf{l \in clause}$}
		         \If{$\mathsf{l \ne outLit\ AND\ marker[\neg l] = 0}$}
		           \State $\mathsf{ok} \leftarrow$ False
		           \Break
		         \EndIf
		      \EndFor
		      \If{$\mathsf{ok\ AND\ (cl, outLit) \not \in gates}$}
		      	\State $\mathsf{gates \leftarrow (cl, outLit)}$
		      \EndIf
		    \EndIf
		  \EndFor

		 \For{$\mathsf{l \in to\_clear}$}
		   \State $\mathsf{marker[l] \leftarrow 0}$
		 \EndFor
	\end{algorithmic}
\end{algorithm}

\subsection{The $\mathsf{solve}$ Algorithm of MiniSat}
\label{appendix:cdcl-algo}
The pseudocode sketch for ${\solve}\mathsf{(assumps)}$ function is shown in Algorithm~\ref{alg:cdcl-solve}, presented in the style of the seminal MiniSat paper by \cite{ES03}.

The ${\solve}\mathsf{(assumps)}$ algorithm makes use of the following standard data structures and subroutines:
\begin{description}
	\item[$\mathsf{branch\_depth}$] current decision level, incremented every time the subroutine $\mathsf{new\_decision\_level()}$ is called.
	\item[$\mathsf{backtrack\_until(\cdot)}$] subroutine to perform the backtracking to the level given as the argument.
	\item[$\mathsf{pick\_branch\_literal(\cdot)}$] picks an unassigned variable and corresponding value to be assigned. Typically, state of the art solvers use variants of VSIDS~\cite{Chaff01} to pick a branch variable and polarity caching~\cite{10.1007/978-3-540-72788-0_28} to decide the polarity of the literal.
	\item[$\mathsf{value(\cdot)}$] value of the literal assigned currently. If a variable $x$ is assigned True, then $\mathsf{value(x)}$ returns True while $\mathsf{value(\neg x)}$ returns False.
	\item[$\mathsf{analyze\_conflict(\cdot)}$] Performs the  conflict analysis and adds the learnt clause
	\item[$\mathsf{should\_restart()}$] returns True or False based on the underlying restart policy
\end{description}

We now provide a quick overview of {\solve}(\assumps). First observe that there is an outer while loop~\ref{line:solve-outer-begin}--~\ref{line:solve-outer-end} and the algorithm exits either when a satisfying assignment is found or it can conclude that the formula is UNSAT. As noted earlier, we run CDCL loops up to a fixed number of conflicts.
From our perspective, we focus on the lines~\ref{line:solve-assumps-begin}--~\ref{line:solve-assumps-end} wherein the algorithm inserts the assumptions into its decision queue as if it was a decision. Note that subsequent to each insertion of an assumption, a full, until-fixedpoint $\mathsf{propagate()}$ is called, on line~\ref{line:solve-prop}.

\begin{algorithm}[htb]
 	\small
	\caption{$\solve(\assumps)$}
	\label{alg:cdcl-solve}
	\begin{algorithmic}[1]
		\While{True}\label{line:solve-outer-begin}

		\State $\mathsf{branch}$ $\leftarrow$ None
		\For{$\mathsf{i \leftarrow branch\_depth}$; $\mathsf{i < assumps.size()}$ and $\mathsf{branch}$ is None; $\mathsf{i}++$} \label{line:solve-assumps-begin}
		\State $\mathsf{lit \leftarrow assumps[i]}$
		\If{$\mathsf{value(lit)}$ is False}
		\State $\mathsf{backtrack\_until(0);}$ \Return UNSAT
		\EndIf
		\If {$\mathsf{value(lit)}$ is True}
		\State $\mathsf{new\_decision\_level()}$ \Comment{Fake decision level%
		}
		\Continue
		\EndIf
		\State $\mathsf{branch \leftarrow lit}$

		\EndFor \label{line:solve-assumps-end}
		\If{$\mathsf{branch}$ is None}
		\State $\mathsf{branch \leftarrow pick\_branch\_literal()}$
		\If {$\mathsf{branch}$ is None} \Comment{Solution found}
		\State $\mathsf{backtrack\_until(0);}$ \Return SAT
		\EndIf
		\EndIf
		\State $\mathsf{new\_decision\_level()}$
		\State $\mathsf{enqueue(branch)}$
		\State \unskip \texttt{prop:}
		\State $\mathsf{ret \leftarrow propagate()}$ \label{line:solve-prop}
		\If {$\mathsf{ret = conflict}$}
		\State $\mathsf{analyze\_conflict(}$)
		\If {$\mathsf{found\_empty\_conflict()}$}
		\State $\mathsf{backtrack\_until(0)}$;
		\Return UNSAT
		\EndIf
		\State $\mathsf{backtrack\_until(backjmp\_level)}$
		\State \textbf{go to} \texttt{prop}
		\EndIf
		\If {$\mathsf{should\_restart()}$} $\mathsf{backtrack\_until(0)}$
		\EndIf
        \If {$\mathsf{conflict\_limit()}$}
        \State $\mathsf{backtrack\_until(0)}$;
        \Return UNKNOWN
        \EndIf
		\EndWhile\label{line:solve-outer-end}
	\end{algorithmic}
\end{algorithm}


\begin{thebibliography}{10}
\providecommand{\url}[1]{\texttt{#1}}
\providecommand{\urlprefix}{URL }
\providecommand{\doi}[1]{https://doi.org/#1}

\bibitem{BSSMS19}
Baluta, T., Shen, S., Shinde, S., Meel, K.S., Saxena, P.: Quantitative
  verification of neural networks and its security applications. In:
  Proceedings of the 2019 ACM SIGSAC Conference on Computer and Communications
  Security. pp. 1249--1264 (2019)

\bibitem{B56}
Beth, E.W.: On {Padoa}'s method in the theory of definition. In: Journal of
  Symbolic Logic 21. pp. 194--195. No.~2 (1956)

\bibitem{CMV13a}
Chakraborty, S., Meel, K.S., Vardi, M.Y.: A scalable and nearly uniform
  generator of {SAT} witnesses. In: Proc. of CAV. pp. 608--623 (2013)

\bibitem{CMV13b}
Chakraborty, S., Meel, K.S., Vardi, M.Y.: A scalable approximate model counter.
  In: Proc. of CP. pp. 200--216 (2013)

\bibitem{CMV14}
Chakraborty, S., Meel, K.S., Vardi, M.Y.: Balancing scalability and uniformity
  in {SAT} witness generator. In: Proc. of DAC. pp.~1--6 (2014)

\bibitem{de-morgan-algebra}
De~Morgan, A.: Trigonometry and Double Algebra. Taylor, Walton, and Malbery,
  London (1849)

\bibitem{DMPV17}
Duenas-Osorio, L., Meel, K.S., Paredes, R., Vardi, M.Y.: Counting-based
  reliability estimation for power-transmission grids. In: Proc. of AAAI (2017)

\bibitem{ES03}
E{\'e}n, N., S{\"o}rensson, N.: An extensible {SAT}-solver. In: International
  conference on theory and applications of satisfiability testing. pp.
  502--518. Springer (2003)

\bibitem{EGSS14}
Ermon, S., Gomes, C., Sabharwal, A., Selman, B.: Low-density parity constraints
  for hashing-based discrete integration. In: International Conference on
  Machine Learning. pp. 271--279. PMLR (2014)

\bibitem{GHSS07}
Gomes, C.P., Hoffmann, J., Sabharwal, A., Selman, B.: Short {XOR}s for model
  counting: from theory to practice. In: International Conference on Theory and
  Applications of Satisfiability Testing. pp. 100--106. Springer (2007)

\bibitem{IMMV15}
Ivrii, A., Malik, S., Meel, K.S., Vardi, M.Y.: On computing minimal independent
  support and its applications to sampling and counting. Constraints pp. 1--18
  (2015). \doi{10.1007/s10601-015-9204-z}

\bibitem{LM16}
Lagniez, J.M., Lonca, E., Marquis, P.: Improving model counting by leveraging
  definability. In: IJCAI. pp. 751--757 (2016)

\bibitem{LM20}
Lagniez, J.M., Lonca, E., Marquis, P.: Definability for model counting.
  Artificial Intelligence  \textbf{281},  103229 (2020)

\bibitem{LS08}
Liffiton, M.H., Sakallah, K.A.: Algorithms for computing minimal unsatisfiable
  subsets of constraints. Journal of Automated Reasoning  \textbf{40}(1),
  1--33 (2008)

\bibitem{Chaff01}
Malik, S., Zhao, Y., Madigan, C.F., Zhang, L., Moskewicz, M.W.: Chaff:
  Engineering an efficient {SAT} solver. Design Automation Conference pp.
  530--535 (2001).
  \doi{http://doi.ieeecomputersociety.org/10.1109/DAC.2001.935565}

\bibitem{MA20}
Meel, K.S., Akshay, S.: Sparse hashing for scalable approximate model counting:
  theory and practice. In: Proceedings of the 35th Annual ACM/IEEE Symposium on
  Logic in Computer Science. pp. 728--741 (2020)

\bibitem{SM20}
Meel, K.S., Soos, M.: Model counting and uniform sampling instances.
  \url{https://doi.org/10.5281/zenodo.3793090} (May 2020).
  \doi{10.5281/zenodo.3793090}

\bibitem{NavEm05}
Naveh, Y., Emek, R.: {Random Stimuli Generation for Functional Hardware
  Verification as a CP Application}. In: Proc. of CP. Lecture Notes in Computer
  Science, vol.~3709, pp. 882--882. Springer (2005)

\bibitem{DBLP:conf/cp/OstrowskiGMS02}
Ostrowski, R., Gr{\'{e}}goire, {\'{E}}., Mazure, B., Sais, L.: Recovering and
  exploiting structural knowledge from {CNF} formulas. In: Proc. of CP. Lecture
  Notes in Computer Science, vol.~2470, pp. 185--199. Springer (2002)

\bibitem{P01}
Padoa, A.: Essai d'une th{\'e}orie alg{\'e}brique des nombres entiers,
  pr{\'e}c{\'e}d{\'e} d’une introduction logique {\`a} une theorie
  d{\'e}ductive quelconque. In: Biblioth{\`e}que du Congr{\`e}s international
  de philosophie. vol.~3, pp. 309--365 (1901)

\bibitem{10.1007/978-3-540-72788-0_28}
Pipatsrisawat, K., Darwiche, A.: A lightweight component caching scheme for
  satisfiability solvers. In: Marques-Silva, J., Sakallah, K.A. (eds.) SAT
  2007. pp. 294--299. Springer (2007)

\bibitem{gate-recovery-cnf}
Roy, J.A., Markov, I.L., Bertacco, V.: Restoring circuit structure from sat
  instances. In: IWLS. pp. 361--368 (2004)

\bibitem{SE19}
Sashittal, P., El-Kebir, M.: Sharptni: counting and sampling parsimonious
  transmission networks under a weak bottleneck. bioRxiv p. 842237 (2019)

\bibitem{SRSM19}
Sharma, S., Roy, S., Soos, M., Meel, K.S.: Ganak: A scalable probabilistic
  exact model counter. In: IJCAI. vol.~19, pp. 1169--1176 (2019)

\bibitem{S20}
Slivovsky, F.: Interpolation-based semantic gate extraction and its
  applications to {QBF} preprocessing. In: International Conference on Computer
  Aided Verification. pp. 508--528. Springer (2020)

\bibitem{SGM20}
Soos, M., Gocht, S., Meel, K.S.: Tinted, detached, and lazy {CNF-XOR} solving
  and its applications to counting and sampling. In: International Conference
  on Computer Aided Verification. pp. 463--484. Springer (2020)

\bibitem{SM19}
Soos, M., Meel, K.S.: {Bird}: Engineering an efficient {CNF-XOR} sat solver and
  its applications to approximate model counting. In: Proc. of the AAAI (2019)

\bibitem{Valiant79}
Valiant, L.G.: The complexity of enumeration and reliability problems. SIAM
  Journal on Computing  \textbf{8}(3),  410--421 (1979)

\end{thebibliography}
\end{document}